\documentclass[11pt]{article}

\usepackage{acl}
\usepackage{inconsolata}
\usepackage{times}
\usepackage{latexsym}
\usepackage{graphicx}
\usepackage{booktabs}
\usepackage{multirow}
\usepackage{xcolor}
\usepackage{amsmath}
\usepackage[T1]{fontenc}
\usepackage[utf8]{inputenc}
\usepackage{microtype}

\newcommand{\PAIlogo}{\raisebox{3.4pt}{\includegraphics[scale=0.09]{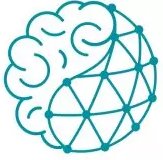}}}
\newcommand{\PAIlogocomma}{\raisebox{3.4pt}{\includegraphics[scale=0.09]{figures/pai-icon.png}\small,}}
\newcommand{\NICElogo}{\raisebox{3.4pt}{\includegraphics[scale=0.010]{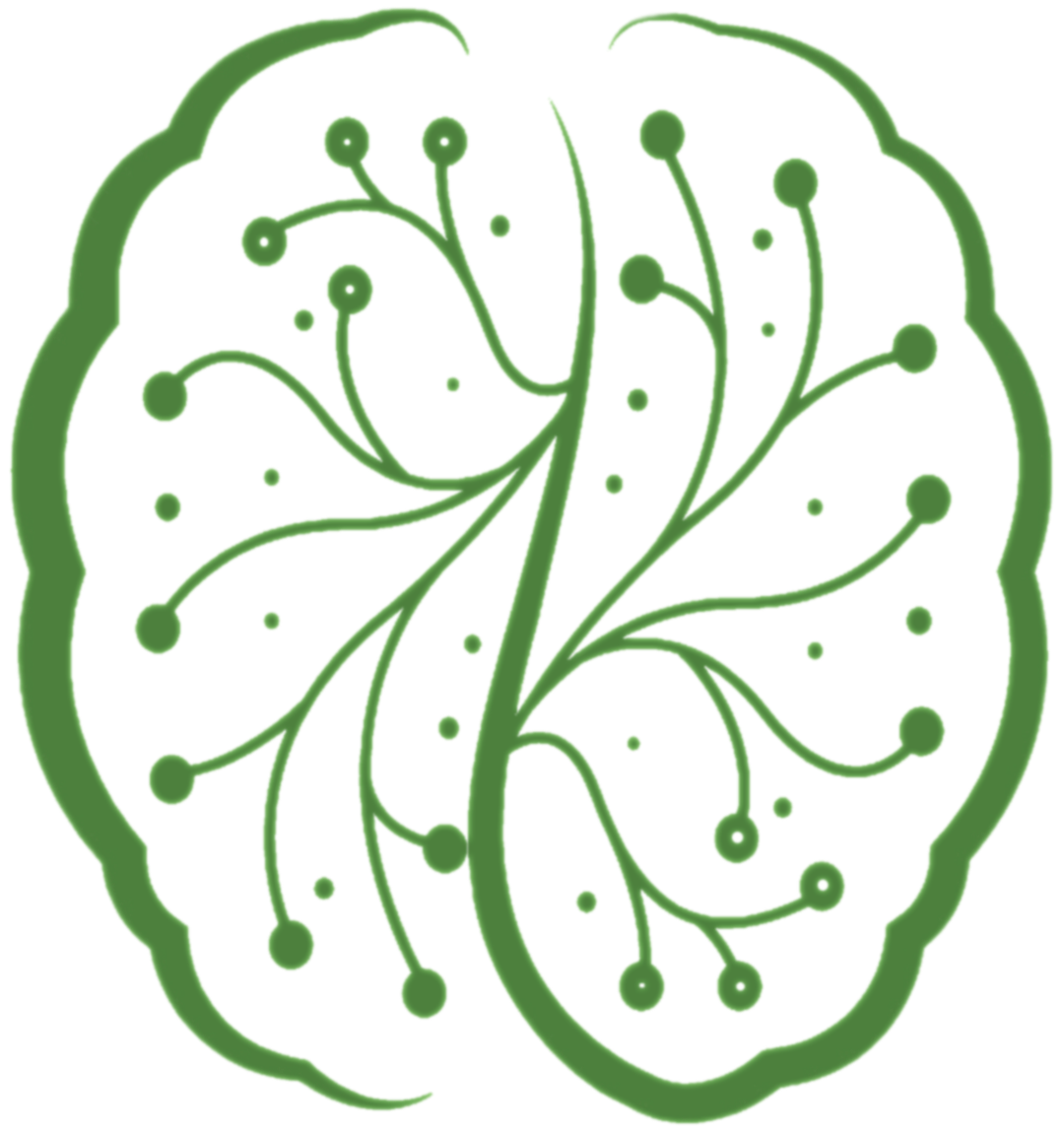}}}
\newcommand{\NICElogocomma}{\raisebox{3.4pt}{\includegraphics[scale=0.010]{figures/Crop-transpBG-ForestGreen-LogoOnly.png}\small,}}

\title{Parameter-Efficient Quality Estimation  \textit{via} Frozen Recursive Models}

\author{
  \textbf{Umar Abubacar}\NICElogo \quad
  \textbf{Roman Bauer}\NICElogocomma\PAIlogo \quad
  \textbf{Diptesh Kanojia}\PAIlogocomma\NICElogo \\[0.5em]
  \NICElogo NICE Research Group, University of Surrey, UK \\
  \PAIlogo Surrey Institute for People-Centred AI (PAI), University of Surrey, UK \\ [0.5em]
  \texttt{\{ua00104, r.bauer, d.kanojia\}@surrey.ac.uk}
}

\begin{document}

\maketitle

\begin{abstract}
Tiny Recursive Models (TRM) achieve strong results on reasoning tasks through iterative refinement of a shared network. We investigate whether these recursive mechanisms transfer to Quality Estimation (QE) for low-resource languages using a three-phase methodology. Experiments on $8$ language pairs on a low-resource QE dataset reveal three findings. First, TRM's recursive mechanisms do not transfer to QE. External iteration hurts performance, and internal recursion offers only narrow benefits. Next, representation quality dominates architectural choices, and lastly, \textit{frozen} pretrained embeddings match fine-tuned performance while reducing trainable parameters by 37$\times$ (7M vs 262M). TRM-QE with frozen XLM-R embeddings achieves a Spearman's correlation of 0.370, matching fine-tuned variants (0.369) and outperforming an equivalent-depth standard transformer (0.336). On Hindi and Tamil, frozen TRM-QE outperforms MonoTransQuest (560M parameters) with 80$\times$ fewer trainable parameters, suggesting that weight sharing combined with frozen embeddings enables parameter efficiency for QE. We release the code publicly for further research\footnote{Code is available at \url{https://github.com/surrey-nlp/TRMQE}}.
\end{abstract}

\section{Introduction}

While the dominant paradigm for enhancing reasoning relies on massive parameter scaling, Tiny Recursive Models (TRM) present a resource-efficient counter-narrative~\citep{jolicoeur2025less}. By applying iterative refinement, \textit{i.e.}, looping data through a shared network multiple times, small models can match the performance of larger single-pass models on reasoning tasks. This raises an intriguing question: can these recursive reasoning mechanisms be transferred to evaluation tasks like Quality Estimation (QE)? 

QE is fundamentally a problem of \textit{cross-lingual reasoning}. Unlike translation, which focuses on generation, QE requires a model to align two semantic spaces, \textit{source} and machine translation output (\textit{hypothesis}), to perform granular error-based reasoning. The model must not only detect deviations but also quantify the severity of them, \textit{i.e.,} numerical reasoning, to predict a quality score. In the current paradigm, large cross-encoders\footnote{multilingual language models} like XLM-R-XXL~\citep{conneau2020unsupervised,goyal-etal-2021-larger} are continually pre-trained, and additionally fine-tuned on large-scale QE data for multiple language pairs, which is expensive for low-resource languages. Given \textit{QE aims to predict translation quality without any reference translations}, it is critical for low-resource languages with limited data.

We investigate whether TRM's recursive efficiency can improve cross-lingual reasoning in resource-constrained scenarios using a QE dataset~\citep{zerva-etal-2022-findings, blain-etal-2023-findings, sindhujan2025alope}  that challenges the model to predict translation quality for English to Indo-Aryan (Hindi, Marathi) and Dravidian (Tamil, Telugu) language families. These translation language pairs involve major typological divergences, such as aligning English's subject–verb–object (SVO) structure with the SOV order of Indic languages, across different scripts. The `reasoning' required here is challenging, as the model must maintain context across long-range dependencies and overcome the relatively poor alignment in pre-trained spaces.

Our investigation demonstrates that while the specific `reasoning' loops of TRM do not directly outperform standard approaches, we observe a ``frozen efficiency'' phenomenon where \textit{frozen} pre-trained embeddings combined with a lightweight, weight-shared TRM head can match the performance of fully fine-tuned baselines while reducing trainable parameters. Our work suggests that, across distinct language families, the key to efficient QE is not adapting the entire large encoder but learning a recursive reasoning mechanism over fixed representations.

\section{Background}

\paragraph{QE as reasoning.} Quality Estimation has evolved from feature-engineered heuristics to deep learning architectures that learn directly from data. Sentence-level QE is typically modelled as a regression task, predicting a score $y$ given a source $s$ and translation $t$. The task is challenging and complex due to the necessary cross-lingual and numerical reasoning. The model must map qualitative errors to a quantitative penalty. For example, a \textit{mistranslated named entity} may incur a higher penalty as compared to a \textit{disfluent} word. Frameworks like TransQuest \citep{ranasinghe2020transquest} tackle this by fine-tuning all parameters of massive encoders. While effective, this couples \textit{representation learning} with \textit{task reasoning}, leading to the compute requirements for fine-tuning and potential overfitting in low-resource settings. Existing work also leverages decoder-based large language models and proposes LoRA-based approaches~\citep{sindhujan2025alope}, and reasoning via constraints like ``annotation guidelines'' within prompts~\citep{sindhujan-etal-2025-llms}.


\paragraph{TRM's architecture} attempts to decouple computational depth from parameter count \citep{jolicoeur2025less}. Unlike standard models where each layer $l$ possesses unique weights $W_l$, TRM utilises a shared transformer block parameterised by $\theta$ applied repeatedly. TRM combines two mechanisms: \textit{internal recursion} through $L$ passes of the shared block (with 2 layers per pass, yielding $2L$ effective layers) and \textit{external iteration} through refinement steps with adaptive halting. On reasoning tasks, small recursive models match larger single-pass models, but this benefit may not transfer to tasks where solutions are directly pattern-matched from inputs. We hypothesise that this recursive structure is relevant for QE as it theoretically allows the model to iteratively re-align source $s$ and translation $t$, verifying semantic consistency akin to human re-reading. We probe if this recurrence can substitute for parameter width, allowing models to `ponder' and refine internal representations before output. TRM achieves strong reasoning performance through a shared transformer applied recursively. We suggest that on reasoning tasks, small recursive models can match larger single-pass models, but this benefit may not transfer to tasks where solutions are directly pattern-matched from inputs.

\section{Experimental Setup}

\subsection{Dataset}

Table~\ref{tab:dataset} summarises the Surrey Low-Resource QE Dataset\footnote{\url{huggingface.co/surrey-nlp/Low-resource-QE-DA-dataset}}, which provides Direct Assessment annotations\footnote{quality score per instance ranging from $0$ to $100$} \citep{specia2018findings,fonseca2019findings} for 8 English-centric language pairs spanning three language families. The dataset totals approximately 75,000 training examples, with most pairs containing 7,000 training and 1,000 test examples. English-Marathi is notably larger with 26,000 training examples.

\begin{table}
  \centering
  
  \begin{tabular}{llll}
    \hline
    \textbf{Pair} & \textbf{Train} & \textbf{Test} & \textbf{Family} \\
    \hline
    en-gu & 7K & 1K & Indo-Aryan \\
    en-hi & 7K & 1K & Indo-Aryan \\
    en-mr & 26K & 699 & Indo-Aryan \\
    en-ta & 7K & 1K & Dravidian \\
    en-te & 7K & 1K & Dravidian \\
    \hline
    et-en & 7K & 1K & Uralic \\
    ne-en & 7K & 1K & Indo-Aryan \\
    si-en & 7K & 1K & Indo-Aryan \\
    \hline
  \end{tabular}
  \caption{Surrey Low-Resource QE Dataset statistics.}
  \label{tab:dataset}
\end{table}

\subsection{Model Architecture}

TRM-QE adapts the TRM architecture for regression by repurposing the adaptive halting output for quality prediction. The original TRM uses a 2-dimensional output head ($q_{halt}$, $q_{continue}$) at the first sequence position for Adaptive Computation Time \citep{graves2016adaptive}. We repurpose sigmoid($q_{halt}$) as our quality score prediction. The model processes concatenated source-translation pairs through pretrained embeddings, applies L-cycles of recursive refinement through a shared transformer block, and produces the final prediction. Our baseline configuration uses 512-dimensional hidden states, 6 L-cycles, 2 layers per cycle, with a 7M parameter transformer core and 255M embedding parameters (262M total). Experiments were conducted on a single NVIDIA A100 GPU, with training taking approximately 5.5 hours.

We evaluate on sentence-level QE, predicting z-normalised Direct Assessment scores transformed to $[0,1]$ via sigmoid to match the model's output layer. Given source-translation pairs without references, we measure performance via Pearson and Spearman correlations.

\begin{figure}[t]
  \centering
  \includegraphics[width=1.0\columnwidth]{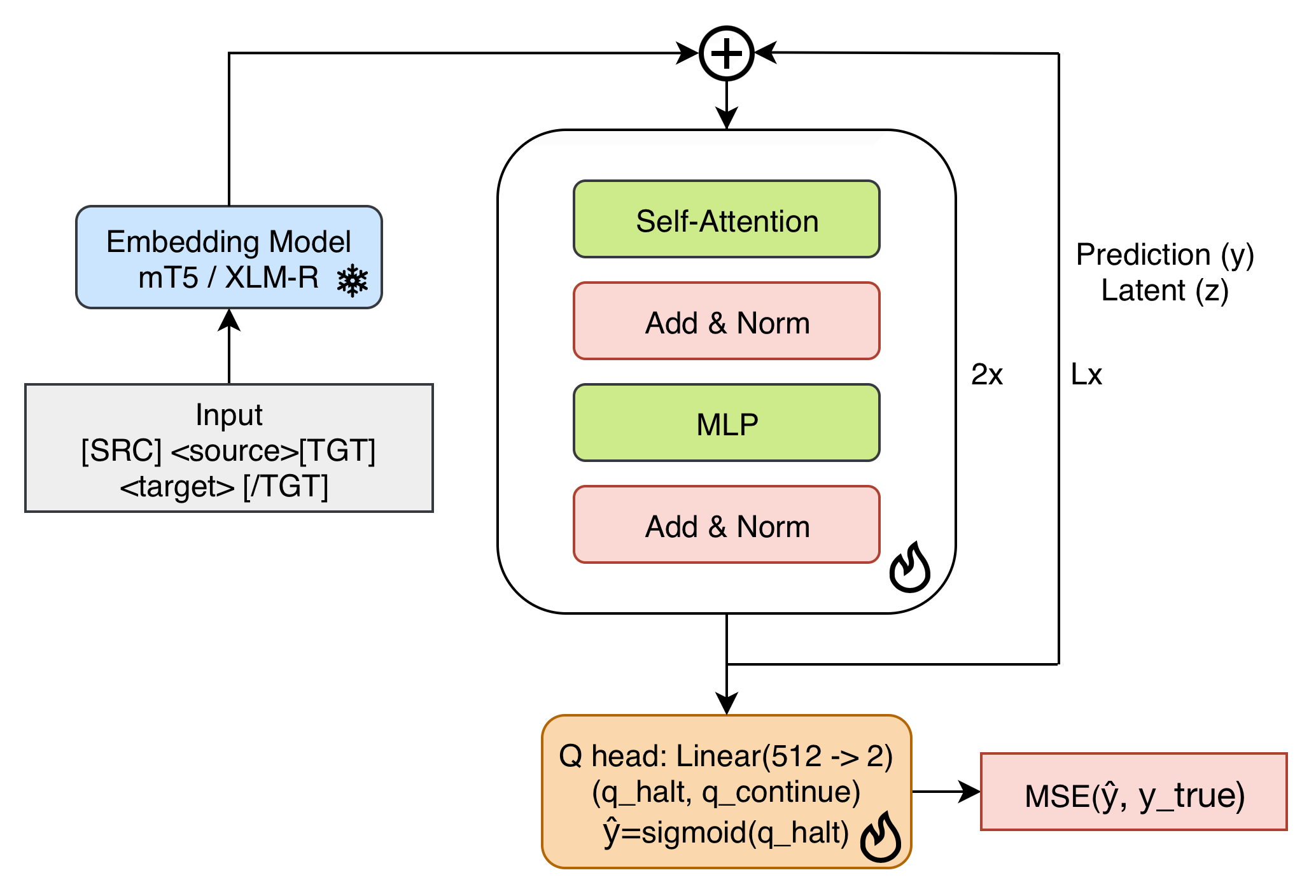}
  \caption{TRM-QE architecture. Source-translation pairs are marked with special tokens and encoded by a pretrained model (frozen or fine-tuned). The TRM applies L weight-shared cycles, then the Q head outputs a quality score via sigmoid activation.}
  \label{fig:architecture}
\end{figure}

\subsection{Three-Phase Methodology}

To disentangle their effects from input representation quality and training efficiency, our work adopts a three-phase methodology testing recursion with fixed representations, varying representations with fixed architecture, and comparing frozen \textit{versus} fine-tuned embeddings.

\textbf{Phase 1} fixes the input representation using fine-tuned mT5-small embeddings (512 dimensions) and systematically varies recursion parameters. We test external iteration steps from 1 to 16 and L-cycles from 1 to 6.

\textbf{Phase 2} fixes the architecture (L=2 cycles, 1 external step) and varies representation quality with fine-tuned embeddings. We compare mT5-small, mT5-base, and XLM-R. For XLM-R (1024-dim), we use SVD projection to 512-dim to maintain consistent model capacity. This phase quantifies how much representation quality matters relative to architectural choices.

\textbf{Phase 3} compares frozen versus fine-tuned embeddings to isolate the contribution of embedding adaptation. By \textit{frozen}, we mean the pretrained XLM-R encoder weights remain fixed during QE training---only the 7M TRM transformer parameters are updated. By \textit{fine-tuned}, we mean all 262M parameters (XLM-R encoder + TRM) are jointly optimised. We also compare fine-tuned TRM against a fine-tuned standard 8-layer transformer without weight sharing to isolate the effect of weight sharing.

\section{Results}

\subsection{Phase 1: Recursion Effects}

\begin{figure}[t]
  \centering
  \includegraphics[width=\columnwidth]{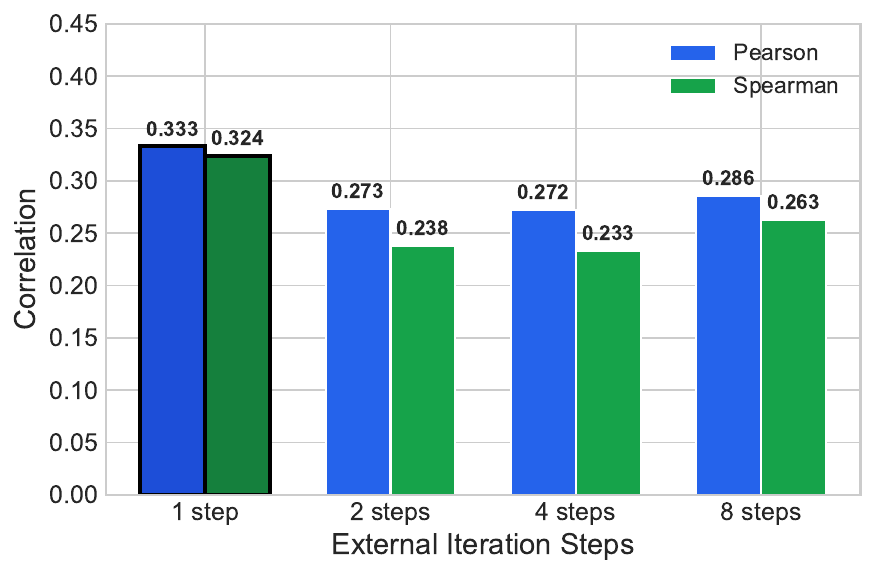}
  \caption{External iteration ablation: single-step models outperform multi-step variants across all step counts tested (1--16).}
  \label{fig:iteration}
\end{figure}

\paragraph{External Iteration} Figure~\ref{fig:iteration} presents results for varying external iteration steps while keeping L-cycles fixed at 6. Contrary to intuitions from reasoning tasks, single-step models achieve the highest correlation (0.333 Pearson, 0.324 Spearman), outperforming multi-step models. Performance degrades sharply from 1 to 2 steps (Spearman drops from 0.324 to 0.238), with no recovery at higher step counts.

\begin{figure}[t]
  \centering
  \includegraphics[width=\columnwidth]{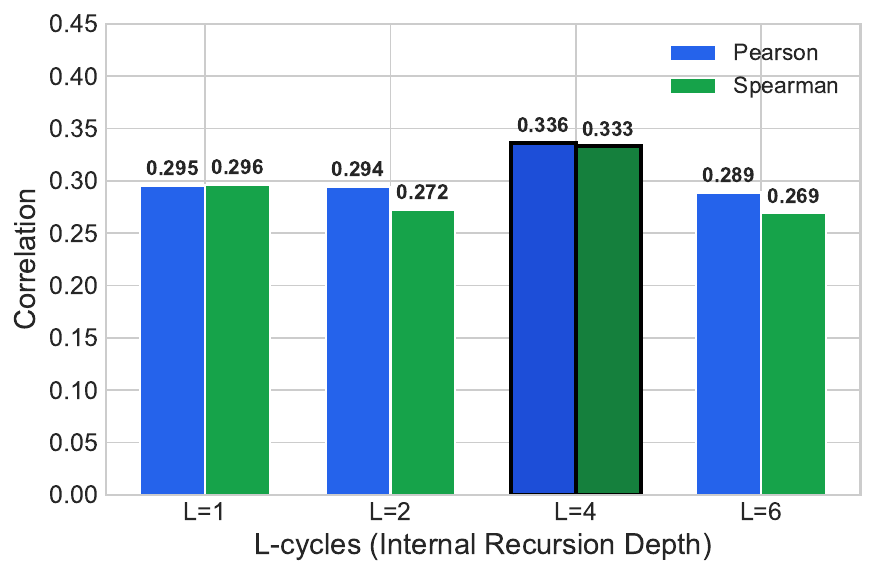}
  \caption{L-cycle ablation: performance peaks at L=4 (8 effective layers), with degradation at both shallower (L=1, L=2) and deeper (L=6) configurations.}
  \label{fig:lcycle}
\end{figure}

\paragraph{Internal Recursion} Figure~\ref{fig:lcycle} shows results for varying L-cycles with external iteration fixed at 1 step. The relationship between depth and performance is non-monotonic. L=1 (2 effective layers) underperforms at 0.295 Pearson, suggesting insufficient capacity for cross-lingual alignment. L=4 (8 layers) achieves peak performance (0.336 Pearson, 0.333 Spearman), while deeper configurations (L=6) degrade. This suggests a sweet spot around L=4 for mT5-small embeddings.

\subsection{Phase 2: Representation Effects}

Using L=2, 1-step configuration, we compare different embedding sources (we revisit optimal L-cycles for XLM-R with frozen embeddings in Phase 3). Table~\ref{tab:embedding-scaling} shows that XLM-R \citep{conneau2020unsupervised} embeddings achieve the best results at 0.387 Pearson and 0.369 Spearman, a 32\% improvement over mT5-small \citep{xue2021mt5} (0.294 Pearson), confirming that representation quality dominates architectural choices for QE.

\begin{table}
  \centering
  \small
  \begin{tabular}{llll}
    \hline
    \textbf{Encoder} & \textbf{Dim} & \textbf{Pearson} & \textbf{Spearman} \\
    \hline
    mT5-small & 512 & 0.294 & 0.272 \\
    \textbf{XLM-R} & \textbf{1024$\to$512} & \textbf{0.387} & \textbf{0.369} \\
    \hline
  \end{tabular}
  \caption{Embedding comparison with L=2, 1-step architecture.}
  \label{tab:embedding-scaling}
\end{table}

\subsection{Phase 3: Frozen vs Fine-tuned Embeddings}

Table~\ref{tab:frozen-comparison} presents an insightful finding: \textit{frozen XLM-R embeddings match or exceed fine-tuned performance while reducing trainable parameters by} 37$\times$. TRM-QE with frozen embeddings (7M trainable) achieves 0.370 Spearman, matching the fine-tuned variant (0.369 with 262M trainable). \textit{However, recursive depth is essential} as frozen L=1 achieves only 0.321 and L=6 degrades to 0.288, confirming L=4 as optimal, in our case. External iteration also hurts frozen models (0.356 with 4 steps vs 0.370 with 1 step), matching the fine-tuned pattern. \textit{Weight sharing is critical for frozen embeddings.} A standard 8-layer transformer with frozen XLM-R achieves only 0.290 Spearman versus TRM's 0.370, a larger gap than for fine-tuned models (0.336 vs 0.369). Frozen embeddings also require strong representations: frozen mT5-small achieves only 0.297 versus frozen XLM-R's 0.370.

\begin{table}
  \centering
  \footnotesize
  \begin{tabular}{@{}lrrr@{}}
    \hline
    \textbf{Model} & \textbf{Train.} & \textbf{Pears.} & \textbf{Spear.} \\
    \hline
    \multicolumn{4}{l}{\textit{Frozen XLM-R (L-cycle ablation)}} \\
    TRM (frz, L=4) & \textbf{7M} & \textbf{0.381} & \textbf{0.370} \\
    TRM (frz, L=1) & 7M & 0.336 & 0.321 \\
    TRM (frz, L=6) & 7M & 0.296 & 0.288 \\
    \hline
    \multicolumn{4}{l}{\textit{Frozen Ablations}} \\
    TRM (frz, L=4, 4 steps) & 7M & 0.369 & 0.356 \\
    Std 8-layer (frz) & 27M & 0.313 & 0.290 \\
    TRM (frz mT5, L=4) & 7M & 0.327 & 0.297 \\
    \hline
    \multicolumn{4}{l}{\textit{Fine-tuned Baselines}} \\
    TRM (ft, L=2) & 262M & 0.387 & 0.369 \\
    Std 8-layer (ft) & 262M & 0.361 & 0.336 \\
    TRM (ft, L=4) & 262M & 0.339 & 0.323 \\
    \hline
  \end{tabular}
   \caption{Frozen vs fine-tuned comparison. frz=frozen, ft=fine-tuned. Frozen XLM-R with TRM (7M) matches fine-tuned (0.370 vs 0.369).}
  \label{tab:frozen-comparison}
\end{table}

The frozen model's success suggests that fine-tuning large embeddings on limited QE data may cause overfitting, while frozen representations combined with TRM's weight-shared architecture provide better generalisation.

\subsection{Comparison with TransQuest}

Tables ~\ref{tab:transquest-comparison} and Table~\ref{tab:per-language-spearman} compare TRM-QE against MonoTransQuest \citep{ranasinghe2020transquest}. While MonoTransQuest achieves higher overall correlation (0.494 vs 0.370), frozen TRM-QE outperforms it on Hindi (+0.097) and Tamil (+0.039) with 80$\times$ fewer trainable parameters.

The performance gap across languages reflects XLM-R's pre-training data distribution \citep{conneau2020unsupervised}: Hindi and Tamil have substantially more Common Crawl data than Telugu and Sinhala. With frozen embeddings, lower-resource languages cannot compensate through task-specific adaptation. \textit{Per-example error analysis confirms this pattern.} Tamil and Hindi achieve both strong correlation (0.556, 0.462) and low mean absolute error (0.098, 0.093) while Telugu predictions are near-random (0.164 Spearman, 0.161 MAE).     

Interestingly, translation direction shows the opposite pattern to prior work: language pairs with English as source (en$\to$X) achieve higher average Spearman (0.405) than those with English as target (X$\to$en, 0.324). This differs from \citet{sindhujan2025alope}, who found English-as-target performed better with fine-tuned models. With frozen embeddings, the model may rely more on source-side features for quality assessment, benefiting when the source is in the better-represented language (English).


\begin{table}
  \centering
  \small
  \begin{tabular}{llll}
    \hline
    \textbf{Model} & \textbf{Trainable} & \textbf{Total} & \textbf{Spearman} \\
    \hline
    TRM-QE (frozen) & \textbf{7M} & 262M & 0.370 \\
    TRM-QE (fine-tuned) & 262M & 262M & 0.369 \\
    MonoTransQuest & 560M & 560M & 0.494 \\
    TransQuest$^\dagger$ & 560M & 560M & 0.592 \\
    \hline
  \end{tabular}
  \caption{Comparison with baselines. $^\dagger$Average from \citet{sindhujan2025alope}. Frozen TRM-QE matches fine-tuned with 37$\times$ fewer trainable parameters.}
  \label{tab:transquest-comparison}
\end{table}

\begin{table}
  \centering
  \resizebox{0.812\columnwidth}{!}{
  \begin{tabular}{lcc}
    \hline
    \textbf{Language} & \textbf{TRM-QE} & \textbf{MonoTQ} \\
    \hline
    en-ta (Tamil) & \textbf{0.556} & 0.517 \\
    en-hi (Hindi) & \textbf{0.462} & 0.365 \\
    en-gu (Gujarati) & 0.423 & \textbf{0.434} \\
    en-mr (Marathi) & 0.418 & \textbf{0.458} \\
    et-en (Estonian) & 0.368 & \textbf{0.741} \\
    ne-en (Nepali) & 0.333 & \textbf{0.593} \\
    si-en (Sinhala) & 0.270 & \textbf{0.527} \\
    en-te (Telugu) & 0.164 & \textbf{0.199} \\
    \hline
    Overall & 0.370 & \textbf{0.494} \\
    \hline
  \end{tabular}
  }
  \caption{Per-language Spearman correlation. TRM-QE (frozen, 7M trainable) vs MonoTransQuest (560M). Bold indicates best per row. TRM-QE outperforms on Tamil and Hindi.}
  \label{tab:per-language-spearman}
\end{table}

\section{Related Work}

TransQuest \citep{ranasinghe2020transquest} established XLM-R fine-tuning as the dominant QE approach. ALOPE \citep{sindhujan2025alope} extends this by adaptively combining Transformer layers from fine-tuned InfoXLM. Our work explores a different direction: frozen pretrained embeddings with a lightweight task head, achieving comparable performance with drastically fewer trainable parameters. TRM \citep{jolicoeur2025less} and Universal Transformers \citep{dehghani2019universal} showed recursive depth can substitute for model size. Concurrent analysis of TRM on ARC-AGI \citep{royeazar2025trm} found that recursion is effectively shallow and most accuracy is achieved at the first step, corroborating our finding that recursion provides limited benefit for QE.

\section{Conclusion}

We find that TRM's recursive mechanisms do not transfer to QE: external iteration hurts performance, internal recursion shows narrow benefits, and representation quality dominates architectural choices. Frozen pretrained embeddings match fine-tuned performance (0.370 vs 0.369 Spearman) while reducing trainable parameters by 37$\times$ (7M vs 262M). However, sufficient recursive depth is essential. Frozen L=1 achieves only 0.321 and L=6 degrades to 0.288, confirming L=4 as optimal. Weight sharing is critical for frozen embeddings as a standard 8-layer transformer with frozen XLM-R achieves only 0.290 versus TRM's 0.370, a larger gap than for fine-tuned models (0.336 vs 0.369).

On Hindi and Tamil, frozen TRM-QE outperforms MonoTransQuest (560M parameters) with 80$\times$ fewer trainable parameters, demonstrating that parameter-efficient QE is viable for language pairs where fine-tuning large models is impractical.

\section*{Limitations}

We evaluate only TRM's recursive architecture on the Surrey Low-Resource dataset; other recursive approaches (Universal Transformers, PonderNet) may behave differently, and validation on WMT QE shared tasks would strengthen generalisation claims. Architectural hyperparameters (L-cycles, external steps) were tuned on mT5-small in Phase 1; optimal settings may differ for other embeddings. Indeed, fine-tuned XLM-R performs best at L=2 rather than L=4. However, the key finding that single-step models outperform multi-step variants holds for both mT5-small and frozen XLM-R. Our adaptation reuses TRM's halting head for quality prediction; we tested a decoupled regression head which showed the same patterns (1-step best) but slightly worse performance. We report bootstrap confidence intervals but not multiple training seeds due to computational constraints; consistent patterns across ablations suggest robust findings. We test only sentence-level QE; word-level or document-level QE may exhibit different recursion dynamics.

\paragraph{Potential Risks.} QE models that underestimate translation errors could propagate low-quality translations in downstream applications. Our frozen approach reduces compute requirements but inherits any biases present in pretrained XLM-R embeddings. Performance varies substantially across languages (Telugu: 0.164 vs Tamil: 0.556), risking unequal benefit across language communities.

\section*{Future Work}

Key directions include investigating why Hindi and Tamil show competitive performance while Telugu and Sinhala lag, testing more permutations of architectures with frozen embeddings and identifying other NLP tasks recursion may benefit.

\section*{Acknowledgments}
This work was supported by the UK Engineering and Physical Sciences Research Council (EPSRC) DTP Studentship 2753922 for the University of Surrey. We thank the Surrey-NLP team for the low-resource QE dataset. Generative AI tools were used to assist with drafting text and code; all content was reviewed by the authors.

\bibliography{references}

@article{jolicoeur2025less,
  author = {Jolicoeur-Martineau, Alexia},
  title = {Less is More: Recursive Reasoning with Tiny Networks},
  journal = {arXiv preprint arXiv:2510.04871},
    year = {2025},
    url = {https://arxiv.org/abs/2510.04871}
}

@inproceedings{sindhujan-etal-2025-llms,
    title = "When {LLM}s Struggle: Reference-less Translation Evaluation for Low-resource Languages",
    author = "Sindhujan, Archchana  and
      Kanojia, Diptesh  and
      Orasan, Constantin  and
      Qian, Shenbin",
    editor = "Hettiarachchi, Hansi  and
      Ranasinghe, Tharindu  and
      Rayson, Paul  and
      Mitkov, Ruslan  and
      Gaber, Mohamed  and
      Premasiri, Damith  and
      Tan, Fiona Anting  and
      Uyangodage, Lasitha",
    booktitle = "Proceedings of the First Workshop on Language Models for Low-Resource Languages",
    month = jan,
    year = "2025",
    address = "Abu Dhabi, United Arab Emirates",
    publisher = "Association for Computational Linguistics",
    url = "https://aclanthology.org/2025.loreslm-1.33/",
    pages = "437--459"
}

@inproceedings{goyal-etal-2021-larger,
    title = "Larger-Scale Transformers for Multilingual Masked Language Modeling",
    author = "Goyal, Naman  and
      Du, Jingfei  and
      Ott, Myle  and
      Anantharaman, Giri  and
      Conneau, Alexis",
    editor = "Rogers, Anna  and
      Calixto, Iacer  and
      Vuli{\'c}, Ivan  and
      Saphra, Naomi  and
      Kassner, Nora  and
      Camburu, Oana-Maria  and
      Bansal, Trapit  and
      Shwartz, Vered",
    booktitle = "Proceedings of the 6th Workshop on Representation Learning for NLP (RepL4NLP-2021)",
    month = aug,
    year = "2021",
    address = "Online",
    publisher = "Association for Computational Linguistics",
    url = "https://aclanthology.org/2021.repl4nlp-1.4/",
    doi = "10.18653/v1/2021.repl4nlp-1.4",
    pages = "29--33"
}

@inproceedings{ranasinghe2020transquest,
  author = {Ranasinghe, Tharindu and Orasan, Constantin and Mitkov, Ruslan},
  title = {{TransQuest}: Translation Quality Estimation with Cross-lingual Transformers},
  booktitle = {Proceedings of the 28th International Conference on Computational Linguistics},
  month = dec,
  year = {2020},
  address = {Barcelona, Spain (Online)},
  publisher = {International Committee on Computational Linguistics},
  pages = {5070--5081},
  doi = {10.18653/v1/2020.coling-main.445}
}

@inproceedings{dehghani2019universal,
  author = {Dehghani, Mostafa and Gouws, Stephan and Vinyals, Oriol and Uszkoreit, Jakob and Kaiser, {\L}ukasz},
  title = {Universal Transformers},
  booktitle = {International Conference on Learning Representations},
  year = {2019},
  url = {https://arxiv.org/abs/1807.03819}
}

@article{graves2016adaptive,
  author = {Graves, Alex},
  title = {Adaptive Computation Time for Recurrent Neural Networks},
  journal = {arXiv preprint arXiv:1603.08983},
  year = {2016},
  url = {https://arxiv.org/abs/1603.08983}
}

@inproceedings{specia2018findings,
  author = {Specia, Lucia and Blain, Fr{\'e}d{\'e}ric and Logacheva, Varvara and Astudillo, Ram{\'o}n F. and Martins, Andr{\'e} F. T.},
  title = {Findings of the {WMT} 2018 Shared Task on Quality Estimation},
  booktitle = {Proceedings of the Third Conference on Machine Translation: Shared Task Papers},
  month = oct,
  year = {2018},
  address = {Brussels, Belgium},
  publisher = {Association for Computational Linguistics},
  pages = {689--709},
  doi = {10.18653/v1/W18-6451}
}

@inproceedings{fonseca2019findings,
  author = {Fonseca, Erick and Yankovskaya, Lisa and Martins, Andr{\'e} F. T. and Fishel, Mark and Federmann, Christian},
  title = {Findings of the {WMT} 2019 Shared Tasks on Quality Estimation},
  booktitle = {Proceedings of the Fourth Conference on Machine Translation (Volume 3: Shared Task Papers, Day 2)},
  month = aug,
  year = {2019},
  address = {Florence, Italy},
  publisher = {Association for Computational Linguistics},
  pages = {1--10},
  doi = {10.18653/v1/W19-5401}
}

@inproceedings{conneau2020unsupervised,
  author = {Conneau, Alexis and Khandelwal, Kartikay and Goyal, Naman and Chaudhary, Vishrav and Wenzek, Guillaume and Guzm{\'a}n, Francisco and Grave, Edouard and Ott, Myle and Zettlemoyer, Luke and Stoyanov, Veselin},
  title = {Unsupervised Cross-lingual Representation Learning at Scale},
  booktitle = {Proceedings of the 58th Annual Meeting of the Association for Computational Linguistics},
  month = jul,
  year = {2020},
  address = {Online},
  publisher = {Association for Computational Linguistics},
  pages = {8440--8451},
  doi = {10.18653/v1/2020.acl-main.747}
}

@inproceedings{xue2021mt5,
  author = {Xue, Linting and Constant, Noah and Roberts, Adam and Kale, Mihir and Al-Rfou, Rami and Siddhant, Aditya and Barua, Aditya and Raffel, Colin},
  title = {m{T}5: A Massively Multilingual Pre-trained Text-to-Text Transformer},
  booktitle = {Proceedings of the 2021 Conference of the North American Chapter of the Association for Computational Linguistics: Human Language Technologies},
  month = jun,
  year = {2021},
  address = {Online},
  publisher = {Association for Computational Linguistics},
  pages = {483--498},
  doi = {10.18653/v1/2021.naacl-main.41}
}

@article{sindhujan2025alope,
  author = {Sindhujan, Archchana and Qian, Shenbin and Chan, Chi Chun Matthew and Orasan, Constantin and Kanojia, Diptesh},
  title = {{ALOPE}: Adaptive Layer Optimization for Translation Quality Estimation using Large Language Models},
  journal = {arXiv preprint arXiv:2508.07484},
  year = {2025},
  url = {https://arxiv.org/abs/2508.07484}
}

@inproceedings{zerva-etal-2022-findings,
  author = {Zerva, Chrysoula and Blain, Fr{\'e}d{\'e}ric and Rei, Ricardo and Lertvittayakumjorn, Piyawat and de Souza, Jos{\'e} G. C. and Eger, Steffen and Kanojia, Diptesh and Alves, Duarte and Ora{\c{s}}an, Constantin and Fomicheva, Marina and Martins, Andr{\'e} F. T. and Specia, Lucia},
  title = {Findings of the {WMT} 2022 Shared Task on Quality Estimation},
  booktitle = {Proceedings of the Seventh Conference on Machine Translation (WMT)},
  month = dec,
  year = {2022},
  address = {Abu Dhabi, United Arab Emirates (Hybrid)},
  publisher = {Association for Computational Linguistics},
  pages = {69--99},
  doi = {10.18653/v1/2022.wmt-1.3}
}

@inproceedings{blain-etal-2023-findings,
  author = {Blain, Fr{\'e}d{\'e}ric and Zerva, Chrysoula and Rei, Ricardo and Guerreiro, Nuno M. and Kanojia, Diptesh and de Souza, Jos{\'e} G. C. and Silva, Beatriz and Vaz, T{\^a}nia and Yan, Jingxuan and Azadi, Fatemeh and Ora{\c{s}}an, Constantin and Martins, Andr{\'e}},
  title = {Findings of the {WMT} 2023 Shared Task on Quality Estimation},
  booktitle = {Proceedings of the Eighth Conference on Machine Translation},
  month = dec,
  year = {2023},
  address = {Singapore},
  publisher = {Association for Computational Linguistics},
  pages = {629--653},
  doi = {10.18653/v1/2023.wmt-1.52}
}

@article{royeazar2025trm,
  author = {Roye-Azar, Antonio and Vargas-Naranjo, Santiago and Ghai, Dhruv and Balamurugan, Nithin and Amir, Rayan},
  title = {Tiny Recursive Models on {ARC-AGI-1}: Inductive Biases, Identity Conditioning, and Test-Time Compute},
  journal = {arXiv preprint arXiv:2512.11847},
  year = {2025},
  url = {https://arxiv.org/abs/2512.11847}
}




\end{document}